# Parametric Object Motion from Blur

Jochen Gast    Anita Sellent    Stefan Roth

Department of Computer Science, TU Darmstadt

## Abstract

*Motion blur can adversely affect a number of vision tasks, hence it is generally considered a nuisance. We instead treat motion blur as a useful signal that allows to compute the motion of objects from a single image. Drawing on the success of joint segmentation and parametric motion models in the context of optical flow estimation, we propose a parametric object motion model combined with a segmentation mask to exploit localized, non-uniform motion blur. Our parametric image formation model is differentiable w.r.t. the motion parameters, which enables us to generalize marginal-likelihood techniques from uniform blind deblurring to localized, non-uniform blur. A two-stage pipeline, first in derivative space and then in image space, allows to estimate both parametric object motion as well as a motion segmentation from a single image alone. Our experiments demonstrate its ability to cope with very challenging cases of object motion blur.*

## 1. Introduction

The analysis and removal of image blur has been an active area of research over the last decade [*e.g.*, 5, 11, 17, 34]. Starting with [8], camera shake has been in the focus of this line of work. Conceptually, the blur is treated as a nuisance that should be removed from the image. While the blur needs to be estimated in the form of a blur kernel, its only purpose is to be used for deblurring. In contrast, it is also possible to treat image blur as a signal that allows to recover certain scene properties from the image. One such example is blur from defocus, where the relationship between the local blur strength and depth can be exploited to recover information on the scene depth at each pixel [19]. In this paper, we also treat blur as a useful signal and aim to recover information on the motion in the scene from a single still image. Unlike work dealing with camera shake, which affects the image in a global manner, we consider *localized motion blur* arising from the independent motion of objects in the scene (*e.g.*, the "London eye" in Fig. 1(a)).

Previous work has approached the problem of estimating motion blur as identifying the object motion from a

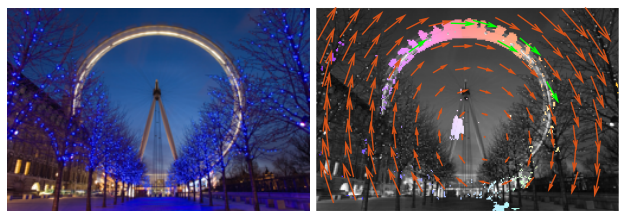

(a) Input image with motion blur    (b) Parametric motion with color-coded motion segmentation

Figure 1. Our algorithm estimates motion parameters and motion segmentation from a single input image.

fixed set of candidate motions [4, 13, 28], or by estimating a non-parametric blur kernel [25] along with the object mask. The former has the problem that the discrete set of candidate blurs restricts the possible motions that can be handled. Estimating non-parametric blur kernels overcomes this problem, but requires restricting the solution space, *e.g.* by assuming spatially invariant motion. Moreover, existing methods are challenged by fast motion, as these require a large set of candidate motions or large kernels, and consequently many parameters, to be estimated. We take a different approach here and are inspired by recent work on optical flow and scene flow, despite the fact that we work with a single input image only. Motion estimation methods have increasingly relied on approaches based on explicit segmentation and parametric motion models [*e.g.* 29, 36, 37] to cope with large motion and insufficient image evidence.

Following that, we propose a parametrized motion blur formulation with an analytical relation between the motion parameters of the object and spatially varying blur kernels. Doing so allows us to exploit well-proven and robust marginal-likelihood approaches [17, 18] for inferring the unknown motion. To address the fact that object motion is confined to a certain region, we rely on an explicit segmentation, which is estimated as part of the variational inference scheme, Fig. 1(b). Since blur is typically estimated in derivative space [8], yet segmentation models are best formulated in image space, we introduce a two-stage pipeline, Fig. 2. First, we estimate the parametric motion along with an initial segmentation in derivative space, and then refine the segmentation in image space by exploiting color models [*e.g.*, 3]. We evaluate our approach on a number of challenging images with significant quantities of localized, non-uniform blur from object motion.

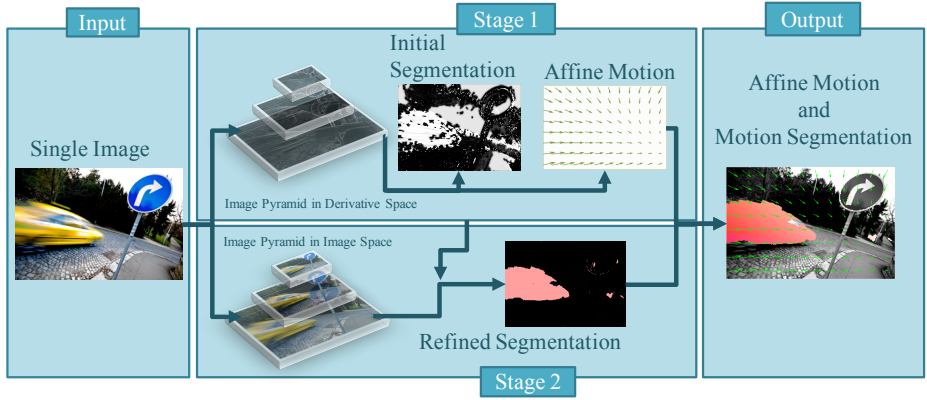

Figure 2. Given a single, locally blurred image as input, our first stage uses variational inference on an image pyramid in derivative space to estimate an initial segmentation and continuous motion parameters, here affine (Sec. 4). Thereby we rely on a parametric, differentiable image formation model (Sec. 3). In a second stage, the segmentation is refined using variational inference on an image pyramid in image space using a color model (Sec. 5). Our final output is the affine motion and a segmentation that indicates where this motion is present.

## 2. Related Work

For our task of estimating parametric motion from a single input image, we leverage the technique of variational inference [12]. Variational inference has been successfully employed in the kernel estimation phase of blind deblurring approaches [8, 21]. As there is an ambiguity between the underlying sharp image and blur kernel estimation, blind deblurring algorithms benefit from a marginalization over the sharp image [17, 18], which we adopt in our motion estimation approach. While it is possible to construct energy minimization algorithms for blind deblurring that avoid these ambiguities [23], this is non-trivial. However, all aforementioned blind deblurring algorithms are restricted to spatially invariant, non-parametric blur kernels.

Recent work lifts this restriction in two ways: First, the space of admissible motions may be limited in some way. To describe blur due to camera shake, Hirsch *et al.* [11] approximate smoothly varying kernels with a basis in kernel space. Whyte *et al.* [31] approximate blur kernels by discretization in the space of 3D camera rotations, while Gupta *et al.* [10] perform a discretization in the space of image plane translations and rotations. Similarly, Zheng *et al.* [38] consider only discretized 3D translations. Using an affine motion model in a variational formulation, our approach does not require discretization of the motion space.

Second, a more realistic description of local object motion may be achieved by segmenting the image into regions of constant motion [4, 16, 25]. To keep the number of parameters manageable, previous approaches either choose the motion of a region from a very restricted set of spatially invariant box filters [4, 16], assume it to have a spatially invariant, non-parametric kernel of limited size [25], or to be discretized in kernel space [13].

Approaches that rely on learning spatially variant blur are similarly limited to a discretized set of detectable motions [6, 28]. Local Fourier or gradient-domain features have been learned to segment motion-blurred or defocused image regions [20, 26]. However, these approaches are designed to be indifferent to the motion causing the blur. Our affine motion model allows for estimating a large variety of practical motions and a corresponding segmentation. In contrast, Kim *et al.* [14] consider continuously varying box filters using TV regularization, but employ no segmentation. However, the problem is highly under-constrained, making it susceptible to noise and model errors.

When given multiple sharp images, the aggregation of smooth motion per pixel into affine motions per layer has a long history in optical flow [*e.g.* 22, 27, 30, 36, 37]. Leveraging motion blur cues for optical flow estimation in sequences affected by blur, Wulff and Black [33] as well as Cho *et al.* [5] use a layered affine model. In an extension of [14], Kim and Lee [15] use several images to estimate motion and sharp frames of a video. In our case of single image motion estimation, Dai and Wu [7] use transparency to estimate affine motion and region segmentation. However, this requires computing local $\alpha$-mattes, a problem that is actually more difficult (as it is more general) than computing parametric object motion. In practice, errors in the $\alpha$-matte and violations of the sharp edge assumption in natural textures lead to inaccurate results. Here we take a more direct approach and consider a generative model of a motion-blurred image, yielding significantly better estimates.

## 3. Parametrized Motion Blur Formation

We begin by considering the image formation in the blurry part of the image, and defer the localization of the blur. Let $\mathbf{y} = (y_i)_i$ be the observed, partially blurred input image, where $i$ denotes the pixel location. Let $\mathbf{x}$ denote the latent sharp image that corresponds to a (hypothetical) infinitesimally short exposure. Since each pixel measures the intensity accumulated over the exposure time $t_f$, we can express the observed intensity at pixel $i$ as the integral

$$y_i = \int_0^{t_f} \mathbf{x}\big(\mathbf{p}_i(t)\big) \, \mathrm{dt} + \epsilon, \quad (1)$$

where $\mathbf{p}_i(t)$ describes which location in the sharp image $\mathbf{x}$ is visible at $y_i$ at a certain time $t$; $\epsilon$ summarizes various noise sources. Note that Eq. (1) assumes that no (dis)occlusion is taking place; violations of this assumption are subsumed in the noise. For short exposures and smooth motion, pixel $y_i$ has only a limited support window $\Omega_i$ in $\mathbf{x}$. Equation (1) can thus be expressed as a spatially variant convolution

$$y_i = \mathbf{k}_i \otimes \mathbf{x}_{\Omega_i} + \epsilon, \quad (2)$$

where the non-uniform blur kernels $\mathbf{k}_i$ hold all contributions from $\mathbf{p}_i(t)$ received during the exposure time. To explicitly construct the blur kernels, we utilize that the motion blur in the blurred part of the image is parametrized by the underlying motion in the scene. Motivated by the fact that rigid motion of planar surfaces can be reasonably approximated by an affine model [1], we choose the parametrization to be a single affine model $\mathbf{u}_i^\mathbf{a}$ with parameters $\mathbf{a} \in \mathbb{R}^6$. Note that other, more expressive parametric models (*e.g.* perspective) are possible. Concretely, we restrict the paths to $\mathbf{p}_i(t) = \big(\frac{t}{t_f} - \frac{1}{2}\big)\mathbf{u}_i^\mathbf{a}$, *i.e.* the integration path depends directly on the pixel location $i$ and affine parameters $\mathbf{a}$, and is constant in time. We now explicitly build continuously valued blur kernels $\mathbf{k}_i^\mathbf{a}$ that allow us to plug the affine motion analytically into Eq. (2).

**Analytical blur kernels.** Given the parametric model we perform discretization in space and time to obtain the kernel

$$\mathbf{k}_i^\mathbf{a}(\boldsymbol{\xi}) = \frac{1}{Z_i^\mathbf{a}} \sum_{t=0}^T \mathrm{psf}\Big(\boldsymbol{\xi} \,\big|\, \big(\tfrac{t}{T} - \tfrac{1}{2}\big)\mathbf{u}_i^\mathbf{a}\Big), \quad (3)$$

where $\boldsymbol{\xi}$ corresponds to the local coordinates in $\Omega_i$ and $Z_i^\mathbf{a}$ is a normalization constant that makes the kernel energy-preserving. $T$ is the number of discretization steps of the exposure interval, and $\mathrm{psf}(\boldsymbol{\xi} \,|\, \boldsymbol{\mu})$ is a smooth, differentiable point-spread function centered at $\boldsymbol{\mu}$ that interpolates the spatial discretization in $\mathbf{x}$. The particular choice of a point-spread function is not crucial, as long as it is differentiable. However, for computational reasons we want the resulting kernels to be sparse. Therefore we choose the point-spread function to be the weight function of Tukey's biweight [9]

$$\mathrm{psf}(\boldsymbol{\xi} \,|\, \boldsymbol{\mu}) = \begin{cases} \Big(1 - \frac{\|\boldsymbol{\xi} - \boldsymbol{\mu}\|^2}{c^2}\Big)^2 & \text{if } \|\boldsymbol{\xi} - \boldsymbol{\mu}\| \leq c \\ 0 & \text{else,} \end{cases} \quad (4)$$

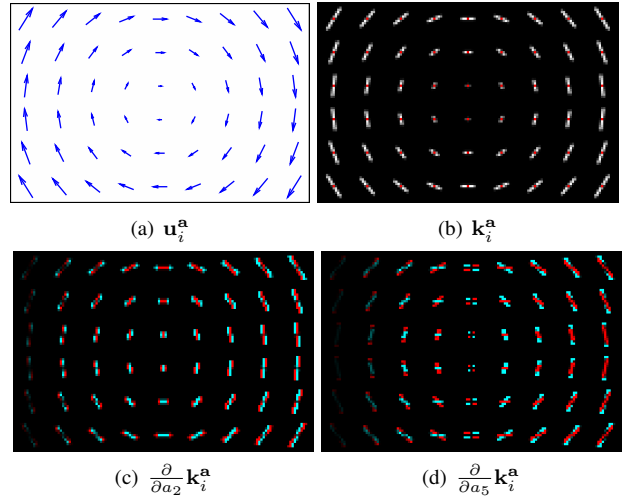

Figure 3. Non-uniform blur kernels at select image locations, and their corresponding derivative filters (positive values – red, negative values – blue) for an example rotational motion. We visualize the derivative filters w.r.t. the rotational parameters $a_2, a_5$. Note how derivative filters change along the $y$-axis for the horizontal component, $a_2$, and the vertical component, $a_5$, respectively.

where $c \in [1, 2]$ controls the width of the constructed blur kernels. For notational convenience, we write the entire image formation process with vectorized images as $\mathbf{y} = \mathbf{K}^\mathbf{a}\mathbf{x}$, where $\mathbf{K}^\mathbf{a}$ denotes a blur matrix holding contributions from all spatially varying kernels $\mathbf{k}_i^\mathbf{a}$ in its rows.

Note that Eq. (3) yields symmetric blur kernels, hence the latent sharp image is assumed to have been taken in the middle of the exposure interval. This is crucial when estimating motion parameters, as it overcomes the directional ambiguity of motion blur. The advantage of an analytical model for the blur kernels is two-fold: First, it allows us to directly map parametrized motion to non-uniform blur kernels, and second, differentiable point-spread functions allow us to compute derivatives w.r.t. the parametrization, *i.e.* $\frac{\partial}{\partial \mathbf{a}}\mathbf{k}_i^\mathbf{a}(\boldsymbol{\xi})$. More precisely, we compute partial derivatives in the form of non-uniform derivative filters acting on the image, *i.e.* $\frac{\partial}{\partial \mathbf{a}}(\mathbf{k}_i^\mathbf{a} \otimes \mathbf{x}_{\Omega_i}) = \big(\frac{\partial}{\partial \mathbf{a}}\mathbf{k}_i^\mathbf{a}\big) \otimes \mathbf{x}_{\Omega_i}$. Figure 3 shows the direct mapping from a motion field to non-uniform blur kernels for various locations inside the image, as well as a selection of the corresponding derivative filters.

**Localized non-uniform motion blur.** Since we are interested in recovering localized object motion rather than global scene motion, the image formation model in Eq. (2) is not sufficient. Here we assume that the image consists of two regions: a static region (we do not deal with camera shake/motion), termed background, and a region that is affected by motion blur, termed foreground. These regions are represented by discrete indicator variables $\mathbf{h} = (h_i)_i$, $h_i \in \{0, 1\}$, which indicate whether a pixel $y_i$ belongs to the blurry foreground. Given the segmentation $\mathbf{h}$, we assume a blurry pixel to be formed as

$$y_i = h_i(\mathbf{k}_i^{\mathbf{a}} \otimes \mathbf{x}_{\Omega_i}) + (1 - h_i)x_i + \epsilon. \quad (5)$$

Although this formulation disregards boundary effects at occlusion boundaries, it has shown good results in the case of constant motion [25]. Note that our generalization to non-uniform parametric blur significantly expands the applicability, but also complicates the optimization w.r.t. the kernel parameters. Despite no closed-form solution, our differentiable kernel parametrization enables efficient inference as we show in the following.

## 4. Marginal-Likelihood Motion Estimation

Estimating any form of motion from a single image is a severely ill-posed problem. Therefore, we rely on a robust probabilistic model and inference scheme. In the related, but simpler problem of uniform blur kernel estimation, marginal-likelihood estimation has proven to be very reliable [18]. We show how more general non-uniform motion models can be incorporated into marginal-likelihood estimation using variational inference. Specifically, we solve for the unknown parametric motion $\mathbf{a}$ while marginalizing over both latent image $\mathbf{x}$ and segmentation $\mathbf{h}$:

$$\hat{\mathbf{a}} = \arg\max_{\mathbf{a}} p(\mathbf{y} \mid \mathbf{a}) \quad (6)$$

$$= \arg\max_{\mathbf{a}} \int p(\mathbf{x}, \mathbf{h}, \mathbf{y} \mid \mathbf{a}) \, d\mathbf{x}\, d\mathbf{h}. \quad (7)$$

We thus look for the point estimate of $\mathbf{a}$ that maximizes the marginal likelihood of the motion parameters. This is enabled by our differentiable blur model from Sec. 3. We model the likelihood of the motion as

$$p(\mathbf{x}, \mathbf{h}, \mathbf{y} \mid \mathbf{a}) = p(\mathbf{y} \mid \mathbf{x}, \mathbf{h}, \mathbf{a})\, p(\mathbf{h})\, p(\mathbf{x}), \quad (8)$$

where we assume the prior over the image, $p(\mathbf{x})$, and the prior over the segmentation, $p(\mathbf{h})$, to factor.

**Likelihood of locally blurred images.** The image formation model (Eq. 5) and the assumption of i.i.d. Gaussian noise $\epsilon$ with variance $\sigma_n^2$ gives rise to the image likelihood

$$p(\mathbf{y} \mid \mathbf{x}, \mathbf{h}, \mathbf{a}) = \prod_i \Big[ \mathcal{N}(y_i \mid \mathbf{k}_i^{\mathbf{a}} \otimes \mathbf{x}_{\Omega_i}, \sigma_n^2)^{h_i} \cdot \\ \mathcal{N}(y_i \mid x_i, \sigma_n^2)^{1-h_i} \Big]. \quad (9)$$

**Segmentation prior.** We assume the object to be spatially coherent and model the segmentation prior with a pairwise Potts model that favors pixels in an 8-neighborhood $N$ to belong to the same segment. Additionally, we favor pixels to be segmented as background if there is insufficient evidence from the image likelihood. We thus obtain

$$p(\mathbf{h}) \propto \prod_i \exp(-\lambda_0 h_i) \cdot \prod_{(i,j)\in N} \exp\big(-\lambda\,[h_i \neq h_j]\big),$$

where $[\cdot]$ is the Iverson bracket and $\lambda, \lambda_0 > 0$ are constants.

**Sharp image prior.** In a marginal-likelihood framework with constant motion, Gaussian scale mixture (GSM) models with $J$ components have been employed successfully [8, 18, 25]. We adopt them here in our framework as

$$p(\mathbf{x}) = \prod_{i,\gamma} \sum_j \pi_j \, \mathcal{N}(f_{i,\gamma}(\mathbf{x}) \mid 0, \sigma_j^2), \quad (10)$$

where $f_{i,\gamma}(\mathbf{x})$ is the $i^\text{th}$ response of the $\gamma^\text{th}$ filter from a set of (derivative) filters $\gamma \in \{1, \ldots, \Gamma\}$ and $(\pi_j, \sigma_j^2)$ correspond to GSM parameters learned from natural image statistics. In log space Eq. (10) is a sum of logarithms, which is difficult to work with. As shown by [18] this issue can be overcome by augmenting the image prior with latent variables, where each variable indicates the scale a particular filter response arises from. Denoting latent indicators for each filter response with $\mathbf{l}_{i,\gamma} = (l_{i,\gamma,j})_j \in \{0,1\}^J$, $\sum_j l_{i,\gamma,j} = 1$, we can write the joint distribution as

$$p(\mathbf{x}, \mathbf{l}) = \prod_{i,\gamma} \prod_j \pi_j^{l_{i,\gamma,j}} \mathcal{N}(f_{i,\gamma}(\mathbf{x}) \mid 0, \sigma_j^2)^{l_{i,\gamma,j}}, \quad (11)$$

where $\mathbf{l}$ is the concatenation of all latent indicator vectors.

### 4.1. Variational inference

Having defined suitable priors and likelihood, we aim to solve Eq. (7) for the unknown motion parameters. Since this problem is intractable, we need to resort to an approximate solution scheme. We use variational approximate inference [18] and define a tractable parametric distribution

$$q(\mathbf{x}, \mathbf{h}, \mathbf{l}) = q(\mathbf{x}) q(\mathbf{h}) \prod_{i,\gamma} q(\mathbf{l}_{i,\gamma}), \quad (12)$$

where we assume the approximating image distribution to be Gaussian with diagonal covariance $q(\mathbf{x}) = \mathcal{N}(\mathbf{x} \mid \mu_\mathbf{x}, \text{diag}(\boldsymbol{\sigma}_\mathbf{x}))$ [18]. The approximate segmentation distribution is assumed to be pixel-wise independent Bernoulli $q(\mathbf{h}) = \prod_i r_i^{h_i}(1-r_i)^{1-h_i}$ and the approximate indicator distribution to be multinomial $q(\mathbf{l}_{i,\gamma}) = \prod_j v_{i,\gamma,j}^{l_{i,\gamma,j}}$, s.t. $\sum_j v_{i,\gamma,j} = 1$.

**Variational free energy.** In the marginal-likelihood framework we directly minimize the KL-divergence between the approximate distribution and the augmented motion likelihood $\text{KL}(q(\mathbf{x}, \mathbf{h}, \mathbf{l}) \parallel p(\mathbf{x}, \mathbf{h}, \mathbf{l}, \mathbf{y} \mid \mathbf{a}))$ w.r.t. the parameters of $q(\mathbf{x}, \mathbf{h}, \mathbf{l})$ and the unknown affine motion $\mathbf{a}$. Doing so maximizes a lower bound for the term $p(\mathbf{y} \mid \mathbf{a})$ in Eq. (6) [18]. The resulting free energy decomposes into the expected augmented motion likelihood and an entropy term

$$F(q, \mathbf{a}) = -\int q(\mathbf{x}, \mathbf{h}, \mathbf{l}) \log p(\mathbf{x}, \mathbf{h}, \mathbf{l}, \mathbf{y} \mid \mathbf{a}) \, d\mathbf{x}\, d\mathbf{h}\, d\mathbf{l} \\ + \int q(\mathbf{x}, \mathbf{h}, \mathbf{l}) \log q(\mathbf{x}, \mathbf{h}, \mathbf{l}) \, d\mathbf{x}\, d\mathbf{h}\, d\mathbf{l}, \quad (13)$$

which we want to minimize. Relegating a more detailed derivation to the supplemental material, the free energy works out as

$$F(q, \mathbf{a}) = \int q(\mathbf{x})q(\mathbf{h}) \frac{\|\mathbf{h} \circ (\mathbf{K}^{\mathbf{a}}\mathbf{x}) + (\mathbf{1}-\mathbf{h}) \circ \mathbf{x} - \mathbf{y}\|^2}{2\sigma_n^2} \, d\mathbf{x} \, d\mathbf{h}$$
$$+ \int q(\mathbf{x}) \big( \sum_{i,\gamma,j} v_{i,\gamma,j} \frac{\|f_{i,\gamma}(\mathbf{x})\|^2}{2\sigma_j^2} \big) \, d\mathbf{x}$$
$$+ \sum_{i,\gamma,j} v_{i,\gamma,j} (\log \sigma_j - \log \pi_j + \log v_{i,\gamma,j})$$
$$+ \lambda_0 \sum_i r_i + \lambda \sum_{(i,j) \in N} r_i + r_j - 2r_i r_j$$
$$+ \sum_i r_i \log r_i + (1 - r_i) \log(1 - r_i)$$
$$- \tfrac{1}{2} \sum_i \log(\boldsymbol{\sigma}_\mathbf{x})_i + \text{const}. \quad (14)$$

Minimizing this energy w.r.t. $q$ and $\mathbf{a}$ is not trivial, as various variables occur in highly non-linear ways. Note that the non-linearities involving the blur matrix $\mathbf{K}^{\mathbf{a}}$ do not occur in previous variational frameworks, where blur kernels are directly considered as unknowns themselves [8, 18, 21] or are part of a discretized representation linear in the unknowns [31], essentially rendering the kernel update into a quadratic programming problem. Despite these issues, we show that one can still minimize the free energy efficiently. More precisely, we employ a standard coordinate descent; *i.e.* at each update step one set of parameters is optimized, while the others are held fixed. In each update step we make use of a different optimization approach to accommodate the dependencies on this particular parameter best.

**Image and segmentation update.** Since we employ the same image prior as previous work [8, 17, 25], the alternating minimization of $F(q, \mathbf{a})$ w.r.t. $q(\mathbf{x})$ and $q(\mathbf{l})$ is similar and can be done in closed form (see supplemental). For updating the segmentation, we have to use a different scheme. Isolating the terms for $\mathbf{r}$ we obtain

$$F(q, \mathbf{a}) = \mathbf{g}(q(\mathbf{x}), \mathbf{a}, \mathbf{y})^T \mathbf{r} + \lambda \sum_{(i,j) \in N} r_i + r_j - 2r_i r_j$$
$$+ \sum_i r_i \log r_i + (1 - r_i) \log(1 - r_i) + \text{const},$$

where $\mathbf{g}(q(\mathbf{x}), \mathbf{a}, \mathbf{y})^T$ models the unary contributions from the expected likelihood and the segmentation prior in Eq. (14). The energy is non-linear due to the entropy terms as well as the quadratic terms in the Potts model. However, the segmentation update does not need to be optimal and it is sufficient to reduce the free energy iteratively. We thus use variational message passing [32], interchanging messages whenever we update the segmentation according to

$$\mathbf{r}_{\text{new}} = \sigma\big( -\mathbf{g}(q(\mathbf{x}), \mathbf{a}, \mathbf{y}) - \lambda L_N \mathbf{1} + 2\lambda L_N \mathbf{r}_{\text{old}} \big), \quad (15)$$

where $\sigma$ is the sigmoid function, and $L_N$ an adjacency matrix for the 8-neighborhood.

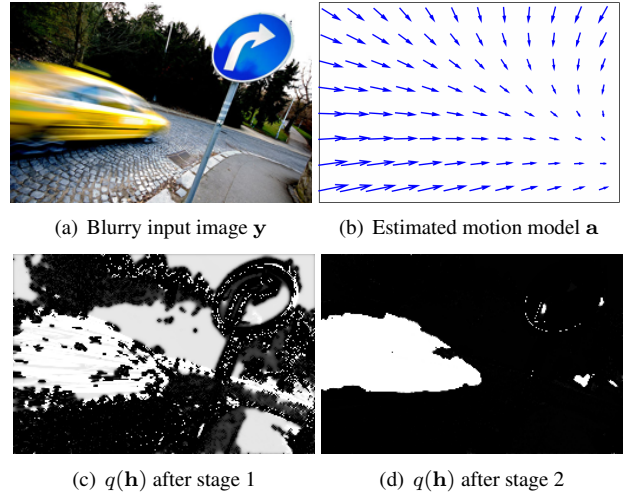

(a) Blurry input image $\mathbf{y}$    (b) Estimated motion model $\mathbf{a}$

(c) $q(\mathbf{h})$ after stage 1    (d) $q(\mathbf{h})$ after stage 2

Figure 4. Improved segmentation after inference in image space.

**Motion estimation.** The unknown motion occurs as a parameter in the expected likelihood. We could deploy a black-box optimizer, however this is very slow. A far more efficient approach is to leverage the derivatives of the analytic blur model and linearize the blur matrix $\mathbf{K}^{\mathbf{a}}$ for small deviations from the current motion estimate $\mathbf{a}^0$ as

$$\mathbf{K}^{\mathbf{a}} \approx \mathbf{K}^0 + \sum_{p=1}^{6} \frac{\partial \mathbf{K}^0}{\partial a_p} d_p =: \mathbf{K}^{\mathbf{d}}, \quad (16)$$

where $\mathbf{d} = \mathbf{a} - \mathbf{a}^0$ is an unknown increment vector.

We locally approximate $\mathbf{K}^{\mathbf{a}}$ by $\mathbf{K}^{\mathbf{d}}$ in Eq. (14) and minimize Eq. (14) w.r.t. subsequent increments $\mathbf{d}$. This is essentially a non-linear least squares problem with an additional term. In the supplemental material we show how the additional term can be linearized and our approach can thus profit from non-linear least-squares algorithms that are considerably faster than black box optimizers.

## 5. Two-Stage Inference and Implementation

To speed up the variational inference scheme and increase its robustness, we employ several well-known details. First, we perform coarse-to-fine estimation on an image pyramid (scale factor 0.5), which speeds up computation and avoids those local minima that present themselves only at finer resolutions [8]. Second, we work in the image gradient domain [18]. Theoretically speaking, the formation model is only valid in the gradient domain for spatially invariant motion, but practically, the step sizes used in our optimization are sufficiently small. The key benefit of the gradient domain is that the variational approximation with independent Gaussians is more appropriate [18].

While motion parameters can be estimated well in this way (*c.f.* Figs. 4(b) and 5), segmentations for regions without significant structure are quite noisy and may contain holes (Fig. 4(c)). This is due to the inherent ambiguity between texture and blur. Smooth regions belong either to a textured, fast-moving object, or an untextured static object. While we cannot always resolve this ambiguity in textureless areas, it thus also does not mislead motion estimation.

To refine the segmentation, we propose a two-stage approach; see Fig. 2 for an overview. In the first stage, we work in the gradient domain and obtain an estimate for the affine motion parameters and an initial, noisy segmentation. In a second stage, we work directly in the image domain. We keep the estimated motion parameters fixed and initialize the inference with the segmentations from the first stage. Moreover, we augment the segmentation prior from Sec. 4 with a color model [3] based on Gaussian mixtures for both foreground and background:

$$\tilde{p}(\mathbf{h} \,|\, \theta_f, \theta_b) \propto p(\mathbf{h}) \\ \cdot \Big[ \prod_i \text{GMM}(y_i \,|\, \theta_f)^{h_i} \text{GMM}(y_i \,|\, \theta_b)^{1-h_i} \Big]^{\lambda_c}. \quad (17)$$

Here, $\theta_f, \theta_b$ are the color distributions and $\lambda_c$ is a weight controlling the influence of the color statistics. We alternate between updating the segmentation and the color statistics of the foreground/background. Empirically, we find that exploiting color statistics in the image space, which is not possible in the gradient domain, significantly improves the accuracy of the motion segmentation (see Fig. 4(d)).

**Initialization.** Since our objective is non-convex, results depend on the initialization on the coarsest level of the image pyramid. We initialize the segmentation with $q(h_i) = 0.5$, *i.e.* foreground and background are equally likely initially. Interestingly, we cannot initialize the motion with $\mathbf{a} = \mathbf{0}$, as the system matrix in the motion update then becomes singular. This is due to the impossibility of determining the "arrow of time" [24], as Eq. (5) is symmetric w.r.t. the sign of $\mathbf{a}$. Since the sign is thus arbitrary, we initialize $\mathbf{a}$ with a small, positive translatory motion. We analyze the sensitivity of our approach to initialization in the supplemental material, revealing that our algorithm yields consistent results across a wide range of starting values.

## 6. Experiments

**Quantitative evaluation.** We synthetically generated 32 test images that contain uniform linear object motion or non-uniform affine motion. For these images we evaluate the segmentation accuracy with the intersection-over-union (IoU) error and motion estimation with the average endpoint error (AEP). While we address the more general non-uniform case, we compare to state-of-the-art object motion estimation approaches that in their public implementations consider only linear motion [4, 7]. Table 1 shows the quantitative results; visual examples can be found in the supplemental material. Our approach performs similar to [4] in the more restricted uniform motion case, but shows a considerably better performance than [4] for the more general non-uniform motion. We can thus address a more general setting without a large penalty for simpler uniform object motion. The method of [7] turns out not to be competitive even in the uniform motion case.

Table 1. Quantitative evaluation.

| Method | segmentation score (IoU) | | motion error (AEP) | |
|---|---|---|---|---|
| | uniform | non-uniform | uniform | non-uniform |
| [4] | **0.53** | 0.33 | **3.84** | 13.37 |
| [7] | – | – | 17.21 | 15.06 |
| Ours | 0.50 | **0.43** | 4.81 | **7.43** |

**Qualitative evaluation.** We conduct qualitative experiments on motion-blurred images that contain a single blurry object under non-uniform motion blur. Such images frequently appear in internet photo collections as well as in the image set provided by [26].

Figs. 1 and 5 show results of our algorithm. In addition to the single input image, the figures show an overlay of a grayscale version of the image with the color-coded [2], non-constant motion field within the segmented region. The global affine motion is additionally visualized with an arrow plot, where green arrows show the motion within the moving object and red arrows continue the global motion field outside the object. All our visual results indicate the motion with unidirectional arrows to allow grasping the relative orientation, but we emphasize that the sign of the motion vector is arbitrary.

Fig. 1(a) shows a particularly challenging example of motion estimation, as the ferris wheel covers a very small number of pixels. Still, the rotational motion is detected well. Fig. 1(b) shows the results after the first stage of the algorithm. Pixels that are not occluded by the tree branches are segmented successfully already using the Potts prior without the additional color information of the second stage. The scenes in Fig. 5 show that our motion estimation algorithm can clearly deal with large motions that lead to significant blur in the image. Note that large motions can actually help our algorithm, at least in the presence of sufficient texture in the underlying sharp scene, as the unidirectional change of image statistics induced by the motion allows to identify its direction [4]. With small motion, this becomes more difficult and the motion may be estimated less accurately by our approach. Moreover, we note that the continuation of the flow field outside the object fits well with what humans understand to be the 3D motion of the considered object. The last image in Fig. 5 shows that foreground and moving object do not necessarily have to coincide for our approach. While motion estimation is successful even in challenging images, the segmentation results are not always perfectly aligned with the motion-blurred objects. Moreover, holes may appear. There are two reasons for this: First, our image formation model from Eq. (5) is a simple approximation and does not capture transparency effects at boundaries. Second, even the color statistics of the second stage do not always provide sufficient information to correctly segment the motion-blurred objects. For instance, for the yellow car in the 3$^{\text{rd}}$ column, the color statistics of the transparent and textureless windows closely match the statistics of the street and thus give rise to false evidence toward the background. Thus our framework identifies the window as background rather than blurry foreground. Comparing our results to those of the recent learning approach of Sun *et al.* [28] in Fig. 8, we observe that the affine motion is estimated correctly also at the lower right corner, where their discretized motion estimate becomes inaccurate. While our segmentation correctly asserts the shirt of the biker as moving in parallel to the car, the color-based segmentation fails to assign the same motion to the black remainder of the cyclist and the transparent car window. These holes indicate that our algorithm cannot resolve the ambiguity between uniform texture and motion blur in all cases.

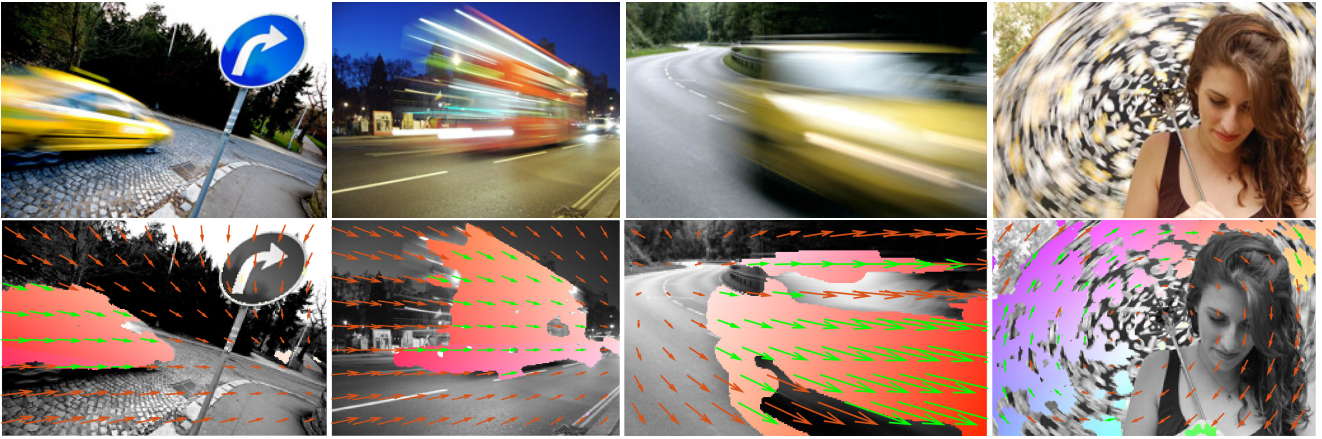

Figure 5. The blurry images (top row) show regions with significant motion blur, as well as static parts. Our algorithm estimates non-uniform motion for all blurry regions and segments out regions dominated by this blur (bottom row, color coding see text).

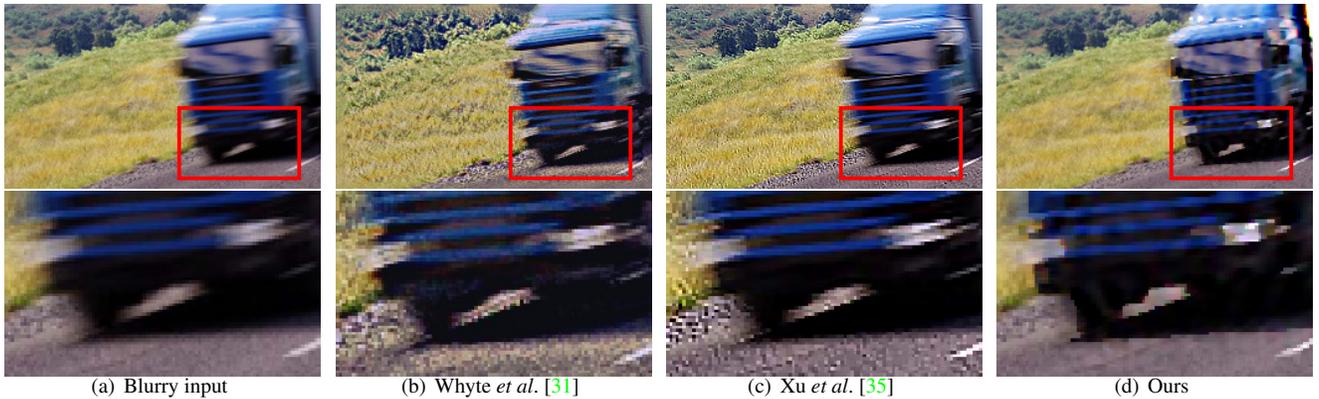

(a) Blurry input  (b) Whyte *et al.* [31]  (c) Xu *et al.* [35]  (d) Ours

Figure 6. Comparison of blind deblurring results. Note the sharpness of the headlights in (d), as well as the over-sharpened road in (b),(c).

We also compare our segmentation results to other blur segmentation and detection algorithms, see Fig. 7. To this end, we have applied the methods of Chakrabarti *et al.* [4] and Shi *et al.* [26]. The results indicate that, despite the discriminative nature of [26], blurry regions and sharp regions are not consistently classified. Fig. 7(c) shows the effect of [4] assuming constant horizontal or vertical motion, thus only succeeds in regions where this assumption is approximately satisfied. Our parametric motion estimation is more flexible and thus identifies blurry objects more reliably.

Although our method is not primarily aimed at deblurring, a latent sharp image is estimated in the course of stage 2. In Fig. 6 we compare this latent sharp image to the reconstructions of [31] and [35]. Using the publicly available implementations, we have chosen the largest possible kernel sizes to accommodate the large blur sizes in our least blurry test image. We observe that our method recovers, *e.g.*, the truck's headlights better than the other methods.

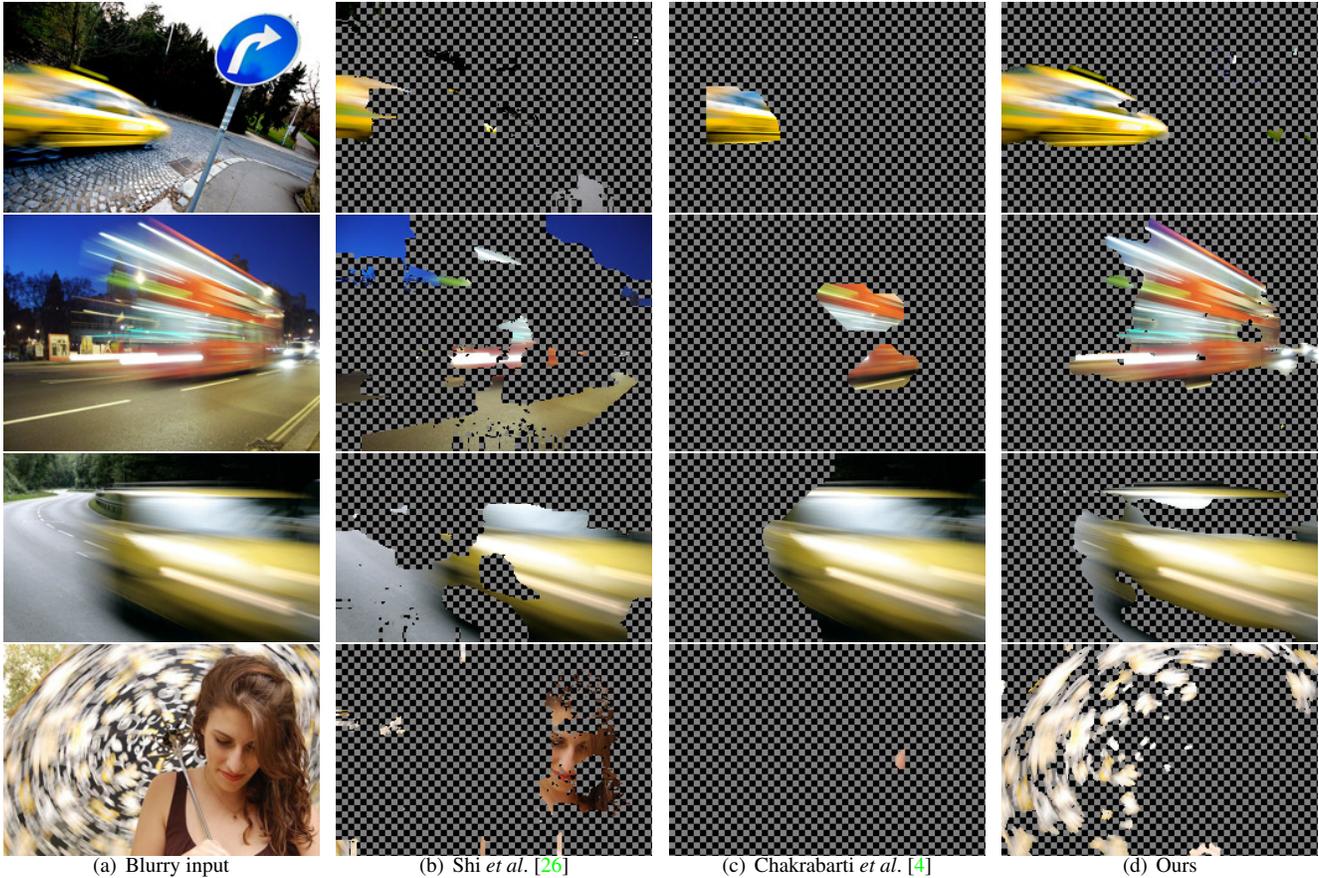

(a) Blurry input  (b) Shi *et al*. [26]  (c) Chakrabarti *et al*. [4]  (d) Ours

Figure 7. For our challenging test scenes (a), blurry-region detection (b) can be mislead by image texture. Generative blur models fare much better (c), as long as the model is sufficiently rich. Our affine motion blur model shows the most accurate motion segmentations (d).

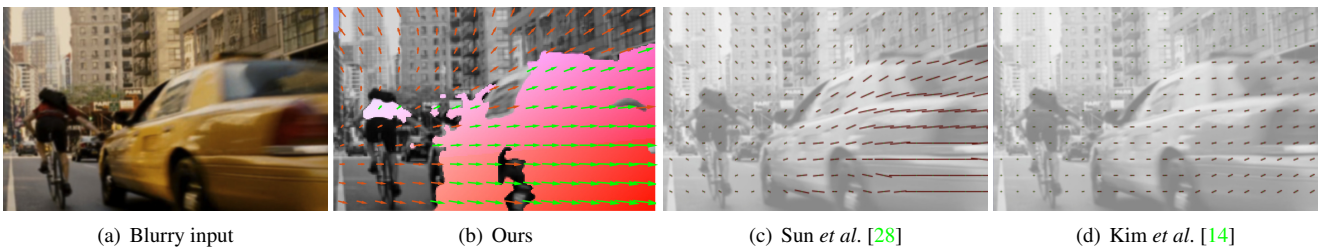

(a) Blurry input  (b) Ours  (c) Sun *et al*. [28]  (d) Kim *et al*. [14]

Figure 8. For the blurry input (a) we compare our approach to two state-of-the art methods (c), (d). Due to its parametric nature our approach recovers the underlying motion of the car more reliably (b). Note that the results of [14, 28] are taken and cropped from [28].

## 7. Conclusion

In this paper we have addressed the challenging problem of estimating localized object motion from a single image. Based on a parametric, differentiable formulation of the image formation process, we have generalized robust variational inference algorithms to allow for the joint estimation of parametric motion and motion segmentation. Our two-stage approach combines the benefits of operating in the gradient and in the image domain: The gradient domain affords accurate variational estimation of the affine object motion and an initial blurry region segmentation. The image domain is then used to refine the segmentation into moving and static regions using color statistics. While our parametric formulation makes variational inference more involved, our results on challenging test data with significant object motion blur show that localized motion can be recovered successfully across a range of settings.

**Acknowledgement.** The research leading to these results has received funding from the European Research Council under the European Union's Seventh Framework Programme (FP7/2007–2013) / ERC Grant Agreement No. 307942.

# Parametric Object Motion from Blur
## – Supplementary Material –

Jochen Gast    Anita Sellent    Stefan Roth
Department of Computer Science, TU Darmstadt

In this supplementary material we show how to derive the free energy as well as the necessary update equations. Moreover, we present additional experimental results.

**Notation.** In the following we will make use of an overloaded notation for both indexing vectors and concatenating scalars (into vectors). That is, whenever a vector $\mathbf{f}$ is given, we retrieve its $i^{\text{th}}$ element via $(\mathbf{f})_i$. On the other hand, we concatenate scalar elements $f_i$ into the vector $(f_i)_i$.

## A. Free Energy

We begin by deriving the free energy as stated in Eq. (14) of the main paper. For conciseness, we express some of the frequently appearing integrals directly as expectations, *i.e.*

$$\int q(\mathbf{x})\phi(\mathbf{x})\,d\mathbf{x} \equiv \langle \phi(\mathbf{x}) \rangle_{q(\mathbf{x})}. \tag{18}$$

Given the independence assumptions of $q(\mathbf{x},\mathbf{h},\mathbf{l})$ in Eq. (12) we can thus rewrite Eq. (13) as

$$\begin{aligned}F(q,\mathbf{a}) = &-\langle \log p(\mathbf{y}\mid\mathbf{x},\mathbf{h},\mathbf{a})\rangle_{q(\mathbf{x},\mathbf{h})} \\ &-\langle \log p(\mathbf{x},\mathbf{l})\rangle_{q(\mathbf{x},\mathbf{l})} - \langle \log p(\mathbf{h})\rangle_{q(\mathbf{h})} \\ &+\langle \log q(\mathbf{x})\rangle_{q(\mathbf{x})} + \langle \log q(\mathbf{h})\rangle_{q(\mathbf{h})} \\ &+\sum_{i,\gamma}\langle \log q(\mathbf{l}_{i,\gamma})\rangle_{q(\mathbf{l}_{i,\gamma})}.\end{aligned} \tag{19}$$

As derived in [8, 18, 25], the entropy terms of the approximating distribution simplify to

$$\langle \log q(\mathbf{x})\rangle_{q(\mathbf{x})} = -\frac{1}{2}\sum_i \log(\boldsymbol{\sigma}_\mathbf{x})_i + \text{const}, \tag{20}$$

$$\langle \log q(\mathbf{h})\rangle_{q(\mathbf{h})} = \sum_i r_i \log r_i + (1-r_i)\log(1-r_i), \tag{21}$$

$$\langle \log q(\mathbf{l}_{i,\gamma})\rangle_{q(\mathbf{l}_{i,\gamma})} = \sum_j v_{i,\gamma,j}\log v_{i,\gamma,j}. \tag{22}$$

The term involving the sparse image prior is similarly derived in [18] and simplifies to

$$\begin{aligned}-\langle \log p(\mathbf{x},\mathbf{l})\rangle_{q(\mathbf{x},\mathbf{l})} = &\left\langle \sum_{i,\gamma,j} v_{i,\gamma,j}\frac{\|f_{i,\gamma}(\mathbf{x})\|^2}{2\sigma_j^2}\right\rangle_{q(\mathbf{x})} \\ &+ \sum_{i,\gamma,j} v_{i,\gamma,j}(\log\sigma_j - \log\pi_j) \\ &+ \text{const}.\end{aligned} \tag{23}$$

To simplify further, we rely on each derivative filter $f_{i,\gamma}$ in Eq. (23) corresponding to a linear operator $\mathbf{D}_\gamma$, *i.e.*

$$f_{i,\gamma}(\mathbf{x}) = (\mathbf{D}_\gamma \mathbf{x})_i. \tag{24}$$

Inserting Eq. (24) into Eq. (23) and expanding the expectation, we can explicitly express it in terms of the moments $(\boldsymbol{\mu}_\mathbf{x}, \boldsymbol{\sigma}_\mathbf{x})$ of $q(\mathbf{x})$:

$$\begin{aligned}-\langle \log p(\mathbf{x},\mathbf{l})\rangle_{q(\mathbf{x},\mathbf{l})} = \\ \sum_{i,\gamma,j} \frac{v_{i,\gamma,j}}{2\sigma_j^2}\bigl(\mathbf{D}_\gamma\boldsymbol{\mu}_\mathbf{x} \circ \mathbf{D}_\gamma\boldsymbol{\mu}_\mathbf{x} + (\mathbf{D}_\gamma \circ \mathbf{D}_\gamma)\boldsymbol{\sigma}_\mathbf{x}\bigr)_i \\ + \sum_{i,\gamma,j} v_{i,\gamma,j}(\log\sigma_j - \log\pi_j) + \text{const}.\end{aligned} \tag{25}$$

where $\circ$ denotes the pointwise Hadamard product.

Using the prior on the segmentation as defined in Sec. 4 of the main paper we obtain

$$\begin{aligned}-\langle \log p(\mathbf{h})\rangle_{q(\mathbf{h})} = \text{const} - &\Bigl\langle \sum_i -\lambda_0 h_i \\ &+ \sum_{(i,j)\in N} -\lambda\,[h_i \neq h_j]\Bigr\rangle_{q(\mathbf{h})} \\ = \lambda_0\sum_i r_i + \lambda\sum_{(i,j)\in N} & r_i + r_j - 2r_i r_j \\ + \text{const}. &\end{aligned} \tag{26}$$

Note that Eq. (26) extends [25] by including the additional bias term $\bigl(\lambda_0 \sum_i r_i\bigr)$ favoring a background segmentation. The last term to be derived involves the expected

i

log-likelihood

$$-\langle \log p(\mathbf{y} \,|\, \mathbf{x}, \mathbf{h}, \mathbf{a})\rangle_{q(\mathbf{x},\mathbf{h})} = \text{const}$$
$$+ \int q(\mathbf{x})q(\mathbf{h}) \frac{\|\mathbf{h} \circ (\mathbf{K}^{\mathbf{a}}\mathbf{x}) + (\mathbf{1}-\mathbf{h}) \circ \mathbf{x} - \mathbf{y}\|^2}{2\sigma_n^2} \, d\mathbf{x} \, d\mathbf{h}, \quad (27)$$

induced by the Gaussian noise assumption of Eq. (9). In order to expand the expectation in Eq. (27), we face the challenge that the latent image $\mathbf{x}$ has a larger domain than the blurry input image $\mathbf{y}$. For this reason we introduce a crop operator $\mathbf{I_y}$ that maps pixel positions of $\mathbf{x}$ to positions in $\mathbf{y}$. Inserting the crop operator and utilizing standard formulas to compute expected values of quadratic norms (see, *e.g.*, [42]) yields the expected log-likelihood

$$-\langle \log p(\mathbf{y} \,|\, \mathbf{x}, \mathbf{h}, \mathbf{a})\rangle_{q(\mathbf{x},\mathbf{h})} = \text{const} \quad (28)$$
$$+ \frac{1}{2\sigma_n^2}\Big(\boldsymbol{\mu}_\mathbf{x}^T \big(\mathbf{K}^{\mathbf{a}T}\mathbf{R}\,\mathbf{K}^{\mathbf{a}} + \mathbf{I_y}^T(\mathbb{I}-\mathbf{R})\,\mathbf{I_y}\big)\boldsymbol{\mu}_\mathbf{x}$$
$$+ \mathbf{r}^T(\mathbf{K}^{\mathbf{a}} \circ \mathbf{K}^{\mathbf{a}})\boldsymbol{\sigma}_\mathbf{x} + (\mathbf{1}-\mathbf{r})^T \mathbf{I_y}\,\boldsymbol{\sigma}_\mathbf{x}$$
$$- 2\boldsymbol{\mu}_\mathbf{x}^T\big(\mathbf{K}^{\mathbf{a}T}\mathbf{R} + \mathbf{I_y}^T(\mathbb{I}-\mathbf{R})\big)\mathbf{y} + \mathbf{y}^T\mathbf{y}\Big),$$

where $\mathbf{R} \equiv \text{diag}(\mathbf{r})$ and $\mathbb{I}$ is the identity matrix. Here, Eq. (28) extends the uniform case in [25] for non-uniform blur matrices $\mathbf{K}^{\mathbf{a}}$.

Inserting all expectations into Eq. (19), we obtain an explicit form of the free energy $F(q, \mathbf{a})$.

## B. Update Equations for Stage 1

Next, we give the update equations w.r.t. the variational parameters of $q$ as well as the motion parameters $\mathbf{a}$ during the first stage (in derivative space).

**Latent indicator update.** Levin *et al.* [18] have shown how to update the GSM (Gaussian scale mixture) indicators $q(\mathbf{l})$ in closed form. Adapting their derivation to our formulation, it is not difficult to see that

$$v_{i,\gamma,j} = \frac{1}{Z_{i,\gamma}} \exp\left(-\frac{1}{2\sigma_j^2}\hat{f}_{i,\gamma}\right)\frac{\pi_j}{\sigma_j}, \quad (29)$$

with

$$Z_{i,\gamma} = \sum_j \exp\left(-\frac{1}{2\sigma_j^2}\hat{f}_{i,\gamma}\right)\frac{\pi_j}{\sigma_j}, \quad (30)$$

$$\hat{f}_{i,\gamma} = \big(\mathbf{D}_\gamma\boldsymbol{\mu}_\mathbf{x} \circ \mathbf{D}_\gamma\boldsymbol{\mu}_\mathbf{x} + (\mathbf{D}_\gamma \circ \mathbf{D}_\gamma)\boldsymbol{\sigma}_\mathbf{x}\big)_i. \quad (31)$$

**Image update.** Isolating the terms involving $\boldsymbol{\mu}_\mathbf{x}$, we obtain the quadratic energy

$$F(q, \mathbf{a}) = \frac{1}{2}\boldsymbol{\mu}_\mathbf{x}^T \mathbf{A}_\mathbf{x} \boldsymbol{\mu}_\mathbf{x} + \mathbf{b}_\mathbf{x}^T \boldsymbol{\mu}_\mathbf{x} + \text{const}, \quad (32)$$

with

$$\mathbf{A}_\mathbf{x} = \frac{1}{\sigma_n^2}\Big(\mathbf{K}^{\mathbf{a}T}\mathbf{R}\,\mathbf{K}^{\mathbf{a}} + \mathbf{I_y}^T(\mathbb{I}-\mathbf{R})\,\mathbf{I_y}\Big)$$
$$+ \sum_{\gamma,j} \frac{1}{\sigma_j^2}\mathbf{D}_\gamma^T \text{diag}(\mathbf{v}_{\gamma,j})\mathbf{D}_\gamma, \quad (33)$$

$$\mathbf{b}_\mathbf{x} = -\frac{1}{\sigma_n^2}\Big(\mathbf{K}^{\mathbf{a}T}\mathbf{R} + \mathbf{I_y}^T(\mathbb{I}-\mathbf{R})\Big)\mathbf{y}, \quad (34)$$

where $\mathbf{v}_{\gamma,j}$ is a vector containing the parameters of the multinomial distribution in the $j^{\text{th}}$ mixture component being associated with the $\gamma^{\text{th}}$ derivative filter. Setting the gradient of Eq. (32) to zero yields a linear system, which can be solved efficiently, *e.g.* using conjugate gradient methods. We can apply similar steps to obtain the update equation for the diagonal covariance $\boldsymbol{\sigma}_\mathbf{x}$, which is given by the element-wise inverse of the diagonal of the linear system for $\boldsymbol{\mu}_\mathbf{x}$.

**Segmentation update.** As explained in the paper, we update the parameters of the Bernoulli distribution of the segmentation by variational message passing (Eq. 15). The required unary contributions are induced by both the bias term in the segmentation (Eq. 26) as well as the expected log-likelihood (Eq. 28):

$$\mathbf{g}(q(\mathbf{x}), \mathbf{a}, \mathbf{y}) = \lambda_0 \mathbf{1} \quad (35)$$
$$+ \frac{1}{2\sigma_n^2}\Big(\mathbf{K}^{\mathbf{a}}\boldsymbol{\mu}_\mathbf{x} \circ \mathbf{K}^{\mathbf{a}}\boldsymbol{\mu}_\mathbf{x} + (\mathbf{K}^{\mathbf{a}} \circ \mathbf{K}^{\mathbf{a}})\boldsymbol{\sigma}_\mathbf{x}$$
$$- 2\,\text{diag}(\mathbf{K}^{\mathbf{a}}\boldsymbol{\mu}_\mathbf{x})\,\mathbf{y}\Big)$$
$$- \frac{1}{2\sigma_n^2}\Big(\mathbf{I_y}\boldsymbol{\mu}_\mathbf{x} \circ \mathbf{I_y}\boldsymbol{\mu}_\mathbf{x} + \mathbf{I_y}\boldsymbol{\sigma}_\mathbf{x} - 2\,\text{diag}(\mathbf{I_y}\boldsymbol{\mu}_\mathbf{x})\,\mathbf{y}\Big),$$

where $\mathbf{1}$ is a vector of all ones.

**Motion update.** We now utilize the parametric nature of our model to efficiently minimize the free energy w.r.t. $\mathbf{a}$. To this end, note that the motion parameters $\mathbf{a}$ exclusively occur in the expected log-likelihood (Eq. 28) and essentially form a quadratic norm plus an additional term accounting for the uncertainty of the latent image. Unfortunately, the parameters $\mathbf{a}$ occur non-linearly within the blur matrix $\mathbf{K}^{\mathbf{a}}$, which makes it hard to obtain a closed-form solution. On the other hand, there are very efficient methods for minimizing non-linear least squares objectives, *i.e.* quadratic norms of non-linear residuals (see [41] for more details). Here, we will adapt such a highly efficient method to our formulation. To begin with, we observe that the motion parameters $\mathbf{a}$ exclusively occur in a subset of terms of the expected log-likelihood

$$F(q, \mathbf{a}) \propto \frac{1}{2}\Big(\boldsymbol{\mu}_\mathbf{x}^T\mathbf{K}^{\mathbf{a}T}\mathbf{R}\,\mathbf{K}^{\mathbf{a}}\boldsymbol{\mu}_\mathbf{x} + \mathbf{r}^T(\mathbf{K}^{\mathbf{a}} \circ \mathbf{K}^{\mathbf{a}})\boldsymbol{\sigma}_\mathbf{x}$$
$$- 2\boldsymbol{\mu}_\mathbf{x}^T\mathbf{K}^{\mathbf{a}T}\mathbf{R}\,\mathbf{y}\Big), \quad (36)$$

where we have dropped constants as well as the factor $\sigma_n^2$, as they are not relevant for the minimization. We continue by linearizing the blur kernels $\mathbf{K^a}$ around the operating point $\mathbf{a}_0$ and express Eq. (36) by means of the linearized blur kernels $\mathbf{K^d}$ with the unknown increment vector $\mathbf{d} \equiv \mathbf{a} - \mathbf{a}_0$ (see Sec. 3 of the paper). Note that linear and quadratic terms involving the non-linear blur matrix can be easily approximated by terms that are linear in $\mathbf{d}$, e.g.

$$\mathbf{K^a x} \approx \mathbf{K^d x} = \mathbf{K^0 x} + \nabla_{\mathbf{a}}(\mathbf{K^0 x})\mathbf{d}, \quad (37)$$

$$\begin{aligned}\mathbf{r}^T(\mathbf{K^a} \circ \mathbf{K^a})\boldsymbol{\sigma_x} &\approx \mathbf{r}^T(\mathbf{K^d} \circ \mathbf{K^d})\boldsymbol{\sigma_x} \\ &= \mathbf{d}^T \mathbf{H}^0(\boldsymbol{\sigma_x})\,\mathbf{d} + 2\,\mathbf{d}^T \mathbf{h}^0(\boldsymbol{\sigma_x}) \\ &\quad + \mathbf{r}^T(\mathbf{K^0} \circ \mathbf{K^0})\boldsymbol{\sigma_x}, \end{aligned} \quad (38)$$

where we define the $N \times 6$ matrix

$$\nabla_{\mathbf{a}}(\mathbf{K^0 x}) = \left(\frac{\partial \mathbf{K^0}}{\partial a_j}\mathbf{x}\right)_j, \quad (39)$$

the $6 \times 6$ matrix

$$\mathbf{H}^0(\boldsymbol{\sigma_x}) = \left(\mathbf{r}^T\left(\frac{\partial \mathbf{K^0}}{\partial a_i} \circ \frac{\partial \mathbf{K^a}}{\partial a_j}\right)\boldsymbol{\sigma_x}\right)_{i,j}, \quad (40)$$

and the $6 \times 1$ vector

$$\mathbf{h}^0(\boldsymbol{\sigma_x}) = \left(\mathbf{r}^T\left(\frac{\partial \mathbf{K^0}}{\partial a_i} \circ \mathbf{K^0}\right)\boldsymbol{\sigma_x}\right)_i. \quad (41)$$

In the expressions above the advantage of the parametric model comes into play as we can efficiently compute $\nabla_{\mathbf{a}}(\mathbf{K^0 x})$, $\mathbf{H}^0(\boldsymbol{\sigma_x})$ and $\mathbf{h}^0(\boldsymbol{\sigma_x})$ by means of the derivative filters $\frac{\partial \mathbf{K^0}}{\partial a_i}$ (see Sec. 4 in the paper). Once we adopt this linearization, the free energy becomes (locally) quadratic in the unknown increment vector $\mathbf{d}$:

$$F(q, \mathbf{d}) = \frac{1}{2}\mathbf{d}^T \mathbf{A}_0\, \mathbf{d} + \mathbf{d}^T \mathbf{b}_0 + \text{const}, \quad (42)$$

with

$$\mathbf{A}_0 = \bigl(\nabla_{\mathbf{a}}(\mathbf{K^0}\boldsymbol{\mu_x})\bigr)^T \mathbf{R}\, \nabla_{\mathbf{a}}\bigl(\mathbf{K^0}\boldsymbol{\mu_x}\bigr) + \mathbf{H}^0(\boldsymbol{\sigma_x}), \quad (43)$$

$$\begin{aligned}\mathbf{b}_0 &= \bigl(\nabla_{\mathbf{a}}(\mathbf{K^0}\boldsymbol{\mu_x})\bigr)^T \mathbf{R}\,\mathbf{K^0}\boldsymbol{\mu_x} + \mathbf{h}^0(\boldsymbol{\sigma_x}) \\ &\quad - \bigl(\nabla_{\mathbf{a}}(\mathbf{K^0}\boldsymbol{\mu_x})\bigr)^T \mathbf{R}\,\mathbf{y}.\end{aligned} \quad (44)$$

We can now use this locally quadratic approximation to minimize the (non-linear) free energy around subsequent operating points. Furthermore, we can build upon regularization techniques from standard non-linear least squares methods, such as the Levenberg-Marquardt approach or use an even more sophisticated step-size control. In our implementation we rely on the Armijo rule [39].

## C. Update Equations for Stage 2

Finally, we give the update equations for the variational parameters during the second stage (in image space).

**Segmentation update.** From Eq. (17) we obtain the free energy

$$\begin{aligned}\tilde{F}(q, \mathbf{a}) &= F(q, \mathbf{a}) - \Bigl\langle \lambda_c \sum_i h_i \log \text{GMM}(y_i \,|\, \theta_f) \\ &\qquad\qquad + (1 - h_i) \log \text{GMM}(y_i \,|\, \theta_b) \Bigr\rangle_{q(\mathbf{h})} \end{aligned} \quad (45)$$

$$\begin{aligned}&= F(q, \mathbf{a}) - \lambda_c \sum_i r_i \log \text{GMM}(y_i \,|\, \theta_f) \\ &\qquad\qquad + (1 - r_i) \log \text{GMM}(y_i \,|\, \theta_b), \end{aligned} \quad (46)$$

augmented by one term accounting for the color statistics of the background/foreground, respectively. In turn, the update for the segmentation in stage 2 differs by one additional unary term:

$$\begin{aligned}\tilde{\mathbf{g}}(q(\mathbf{x}), \mathbf{a}, \mathbf{y}) &= \mathbf{g}(q(\mathbf{x}), \mathbf{a}, \mathbf{y}) \\ &\quad + \lambda_c\bigl(-\log \text{GMM}(y_i \,|\, \theta_f) \\ &\qquad\quad + \log \text{GMM}(y_i \,|\, \theta_b)\bigr)_i. \end{aligned} \quad (47)$$

**Color statistics update.** Let $\theta_f = \{\pi_{f,j}, \mu_{f,j}, \Sigma_{f,j} \,|\, j = 1\ldots J\}$ and $\theta_b = \{\pi_{b,j}, \mu_{b,j}, \Sigma_{b,j} \,|\, j = 1\ldots J\}$ be the parameters of the Gaussian mixture model for the foreground and background colors, respectively. Then Eq. (46) can be minimized by the expectation-maximization (EM) algorithm for Gaussian mixture models, however, each update equation is weighted by the parameters of the Bernoulli distribution of the segmentation.

For instance, updates for the foreground color statistics are given by

$$\alpha_{i,j} = \frac{\pi_j \mathcal{N}(\mathbf{y}_i \,|\, \boldsymbol{\mu}_j, \boldsymbol{\Sigma}_j)}{\sum_k \pi_k \mathcal{N}(\mathbf{y}_i \,|\, \boldsymbol{\mu}_k, \boldsymbol{\Sigma}_k)}, \quad (48)$$

$$N_j = \sum_i r_i \alpha_{i,j}, \quad (49)$$

$$\boldsymbol{\mu}_j^{\text{new}} = \frac{1}{N_j} \sum_i r_i \alpha_{i,j} \mathbf{y}_i, \quad (50)$$

$$\boldsymbol{\Sigma}_j^{\text{new}} = \frac{1}{N_j} \sum_i r_i \alpha_{i,j} (\mathbf{y}_i - \boldsymbol{\mu}_j^{\text{new}})(\mathbf{y}_i - \boldsymbol{\mu}_j^{\text{new}})^T, \quad (51)$$

$$\pi_j^{\text{new}} = N_j / \sum_k N_k, \qu(52)$$

where we dropped the foreground index $f$ for brevity.

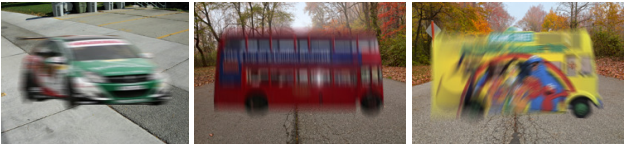

Figure 9. Synthesized uniform and non-uniform motion blur.

## D. Synthetic Dataset

For the quantitative analysis in the paper we created a dataset of 32 test images, divided in two subsets: uniform linear motion and non-uniform affine motion. The test images were created by extracting objects and segmentations from the VOC2012 dataset [40], and pasting them on top of different backgrounds. More precisely, motion blur is simulated by iteratively warping both the extracted images and segmentations according to either uniform or affine motion, and pasting its warped (latent) images on top of the background. While the resulting blurred image is given by the average of all warped latent images, the ground truth segmentation is chosen to be the maximum, *i.e.* the union, of all warped segmentations. Examples are shown in Fig. 9.

## E. Sensitivity Analysis

We analyze the sensitivity of our approach to different initializations. To that end, we created a synthetic example (Fig. 9, left) of horizontal motion ($a_1 = 15$) and measured the resulting average endpoint error of the motion estimation for initializations with increasingly large motion in either vertical or horizontal direction. Table 2 shows the resulting average endpoint errors. Unless our method is initialized with a significant motion in the incorrect (vertical) direction, which leads to a failure (marked red), our algorithm yields consistent results. Initializing with a small motion avoids such issues.

## F. Additional Results

We continue to show a few more results in addition to the ones provided in the main paper. Note that some of these examples are taken from [26].

**Additional examples.** While our approach is primarily aimed at recovering object (foreground) motion, the example in the last row of Fig. 7 of the main paper has already shown that we can also estimate motion and segmentation from a motion blurred background. Figures 10 and 11 show two more such examples in which a sharp bicyclist is shown before a motion-blurred background. Our approach correctly identifies the background scene as the motion-blurred region ("foreground") and vice versa.

Figures 12 and 13 show additional results for a purely rotational ferris wheel, as well as a motion-blurred rollercoaster. Note that the rollercoaster is segmented very well, but the ferris wheel less so. While our variational framework identities the rotational motion correctly, the blurry foreground of the outer wheel blends with the background, hence our approach does not properly pick up these regions as part of the blurry foreground.

**Failure cases.** In Figs. 14 and 15 we show two examples for which the variational framework fails to estimate either a correct motion model or a segmentation. In the first example (Fig. 14) both estimating the motion model as well as estimating the segmentation fails. To be successful, our approach requires a sufficiently large region of observable motion blur; if this is not the case our algorithm may end up in poor local minima. This is in particular the case for estimating the correct motion, since the motion updates are based on iteratively optimizing a non-linear objective by (locally) quadratic approximations. Also observe how the inference heavily picks up the horizontal structures in the background wall on the right-hand side, as they provide evidence for horizontal motion blur.

The second example (Fig. 15) shows how the variational framework fails to estimate the correct motion model due to directional ambiguities. Note how the top left motion vectors point to the left, while a major part of the estimated motion vectors point to the right. In our approach we tackle this ambiguity by explicitly modeling symmetric blur kernels, however, the motion parameters still allow for two equally good explanations: Either the background translates to the left or to the right. In practice we overcome this problem by initializing the motion estimate with a slight bias towards either direction. However, this example indicates that this bias alone may not always be enough to resolve the ambiguities one may observe during the inference process.

Table 2. Average endpoint error after different initializations.

| $|a_{1/4}|$ | 0.1 | 0.5 | 1 | 3 | 5 | 7 |
|---|---|---|---|---|---|---|
| vertical | 0.55 | 0.48 | 0.47 | 17.47 | 21.39 | 28.91 |
| horizontal | 0.53 | 0.56 | 0.51 | 0.57 | 0.52 | 0.53 |

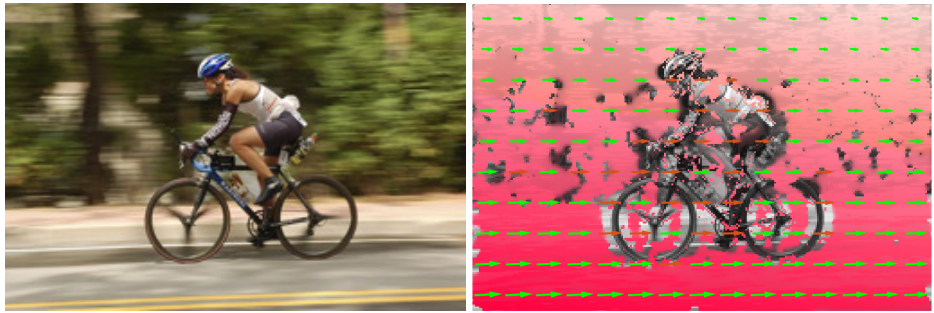

(a) Blurry input  (b) Parametric motion + motion segmentation

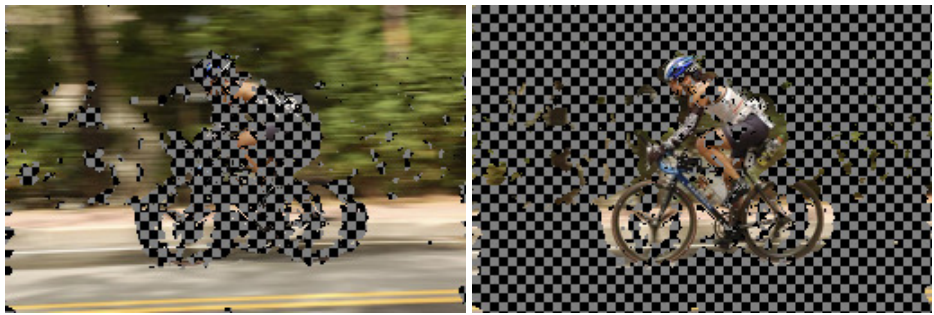

(c) Blurry foreground  (d) Static background

Figure 10. Motion from a blurry background.

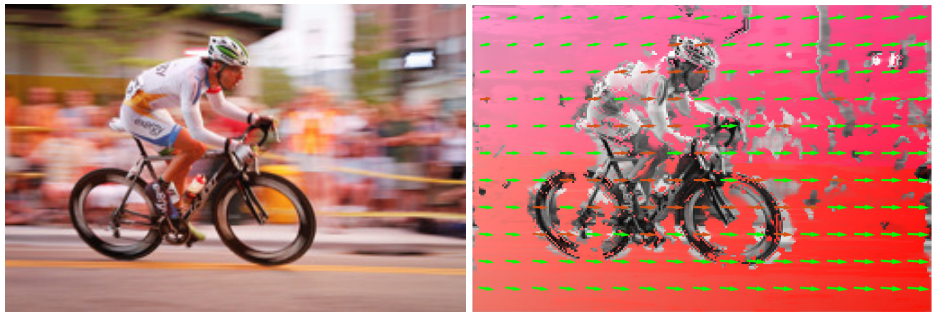

(a) Blurry input  (b) Parametric motion + motion segmentation

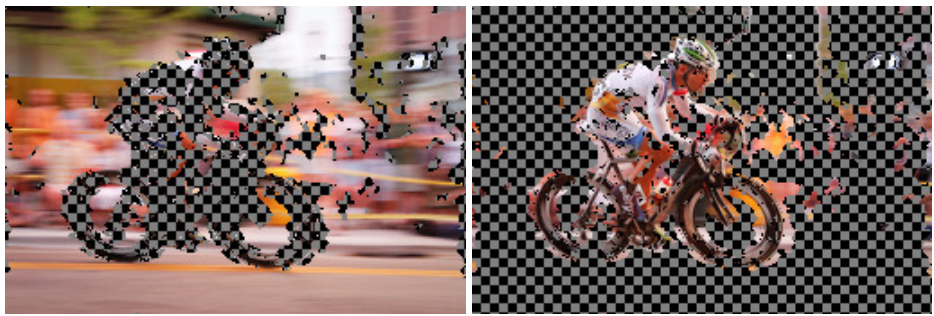

(c) Blurry foreground  (d) Static background

Figure 11. Motion from a blurry background.

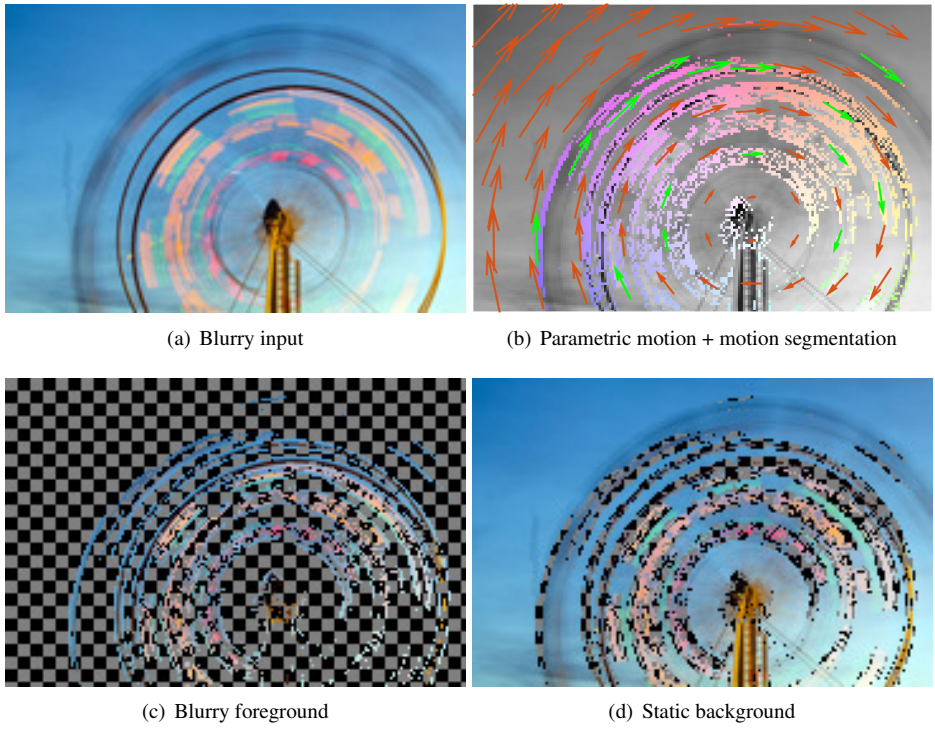

(a) Blurry input     (b) Parametric motion + motion segmentation

(c) Blurry foreground     (d) Static background

Figure 12. Rotational motion from a ferris wheel.

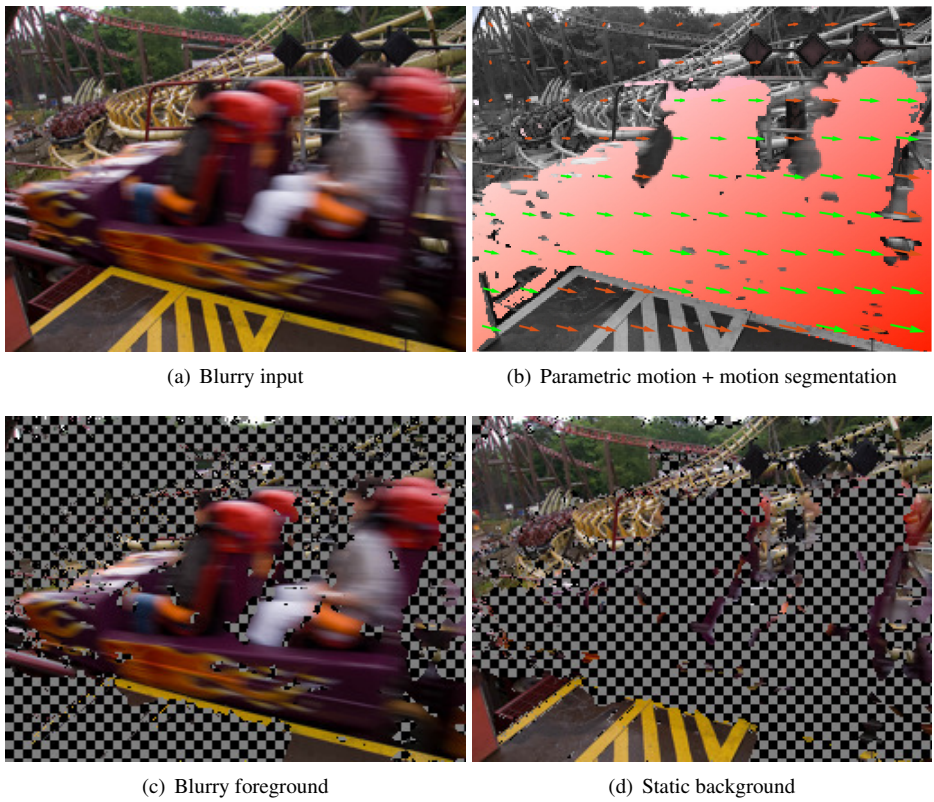

(a) Blurry input     (b) Parametric motion + motion segmentation

(c) Blurry foreground     (d) Static background

Figure 13. Affine motion from a rollercoaster.

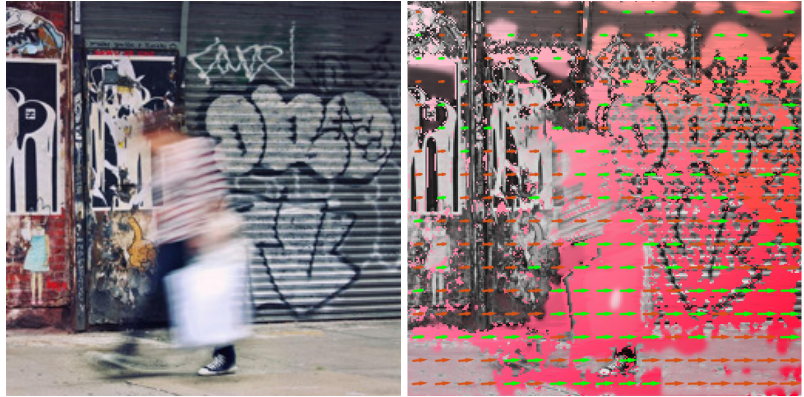

(a) Blurry input      (b) Parametric motion + motion segmentation

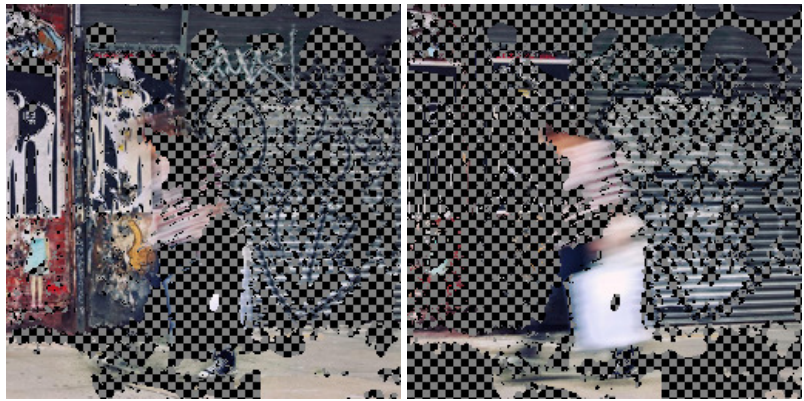

(c) Blurry foreground      (d) Static background

Figure 14. Motion estimation may fail if the blurry region is too small in comparison to the background.

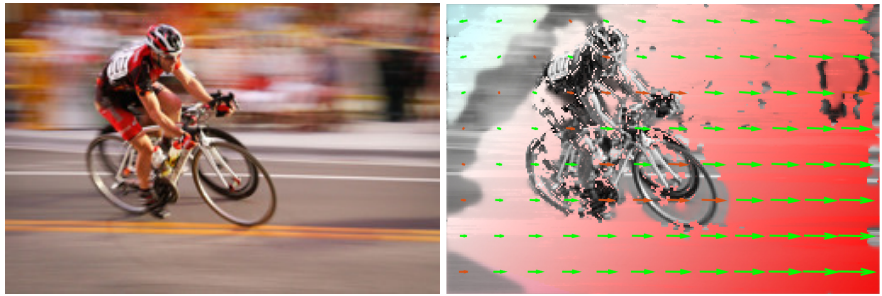

(a) Blurry input      (b) Parametric motion + motion segmentation

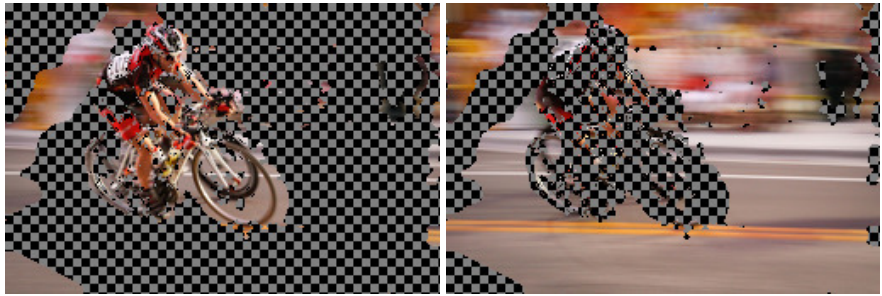

(c) Blurry foreground      (d) Static background

Figure 15. Motion estimation may fail due to directional ambiguities.